\pdfoutput=1

\documentclass[11pt]{article}

\usepackage[]{ACL2023}

\usepackage{times}
\usepackage{latexsym}
\usepackage{algorithm}
\usepackage{arevmath}     
\usepackage[noend]{algpseudocode}
\usepackage{multirow}
\usepackage{supertabular}
\usepackage{graphicx}
\usepackage{caption}
\usepackage{subcaption}

\usepackage[T1]{fontenc}

\usepackage[utf8]{inputenc}

\usepackage{microtype}

\usepackage{inconsolata}

\usepackage{booktabs}
\usepackage{array} 
\usepackage{longtable}
\usepackage{url}

\newcommand{\dataname}{\textsc{FoFo}}
\newcommand{\contentfollowing}{\textsc{Content-Following}}
\newcommand{\formatfollowing}{\textsc{Format-Following}}
\newcommand{\formatinstru}{\textsc{Format-Instru}}
\newcommand{\jiangshu}[1]{{\color{black}{} #1}}
\usepackage{hyperref}

\title{\dataname: A Benchmark to Evaluate LLMs' Format-Following Capability}

\author{
Congying Xia{$^1$\thanks{~ Indicates Equal Contribution}},~Chen Xing{$^1$\footnotemark[1]}, ~Jiangshu Du$^2$, ~Xinyi Yang$^1$, \\ \textbf{~Yihao Feng$^1$, ~Ran Xu$^1$, Wenpeng Yin$^3$, Caiming Xiong$^1$}\\
  {$^1$Salesforce Research}\\
  {$^2$University of Illinois at Chicago}\\
  {$^3$Pennsylvania State University}\\
  {\tt \{c.xia,cxing,x.yang, yihaofeng, ran.xu, cxiong\}@salesforce.com}, \\
  {\tt jdu25@uic.edu},
  {\tt wenpeng@psu.edu}\\
}

\begin{document}
\maketitle
\begin{abstract}
This paper presents {\dataname}, a pioneering benchmark for evaluating large language models' (LLMs) ability to follow complex, domain-specific formats, a crucial yet underexamined capability for their application as AI agents. Despite LLMs' advancements, existing benchmarks fail to assess their format-following proficiency adequately. {\dataname} fills this gap with a diverse range of real-world formats and instructions, developed through an AI-Human collaborative method. Our evaluation across both open-source (e.g., Llama 2, WizardLM) and closed-source (e.g., GPT-4, PALM2, Gemini) LLMs highlights three key findings: open-source models significantly lag behind closed-source ones in format adherence; LLMs' format-following performance is independent of their content generation quality;  and LLMs' format proficiency varies across different domains. These insights suggest the need for specialized tuning for format-following skills and highlight {\dataname}'s role in guiding the selection of domain-specific AI agents. {\dataname} is  released here. \footnote{\url{https://github.com/SalesforceAIResearch/FoFo}}
\end{abstract}

\section{Introduction}
Large language models (LLMs) show great promise in automating diverse tasks, from medical~\citep{thirunavukarasu2023large,shah2023creation, clusmann2023future, tang2023evaluating} and legal data analysis~\citep{cui2023chatlaw,jiang2023legal,fei2023lawbench,guha2023legalbench} to daily assistance with activities like reservations~\citep{ma2023understanding,gao2023s,xi2023rise,muthusamy2023towards}. For LLMs to effectively assist humans, their crucial ability lies in following instructions~\citep{ouyang2022training, chung2022scaling, wang2022self, wei2021finetuned}. In tasks requiring them to act as AI agents, such as organizing medical records or generating KPI reports, precise adherence to specified formats given by humans is essential~\citep{xi2023rise}. Without this capability, the practicality of employing LLMs in such roles diminishes significantly.

However, prior evaluation benchmarks of LLMs fall short in assessing their format-following capabilities.  
On one hand, mainstream instruction-following benchmarks, e.g., AlpacaEval~\citep{dubois2023alpacafarm} and MT-Bench~\citep{zheng2023judging}, are in the open question-answering/chatting style, assessing the correctness of LLM responded content (therefore, we refer to those as \contentfollowing~benchmarks) without explicitly considering their format-following capability.
State-of-the-art LLMs, whether closed or open-source, increasingly demonstrate comparable performance on these \contentfollowing~benchmarks. 
On the other hand, benchmarks evaluating AI agents~\citep{yang2023language,zhou2023webarena,shridhar2020alfworld} in specific environments typically prioritize the overall success rate in completing test tasks.
Since there are many factors that can affect final success rates, such as the model's grounding and reasoning capability in the specific environment, the final success rates cannot directly gauge LLMs' format-following proficiency. Consequently, although we observe LLMs performing notably worse than humans in certain domains~\citep{zhou2023webarena}, it remains unclear if this discrepancy is partly attributed to format-following limitations.  Exploring enhancements in LLMs' format-following may potentially pave the way for further improvements in their role as  AI agents.


In this work, we take the lead to build \dataname, a benchmark specifically designed for assessing LLMs' \formatfollowing~capablities. We equip \dataname~with two shining features: \textit{a wide coverage of real-world domain-specific formats (such as HL7-CDA format~\citep{dolin2001hl7} in Healthcare) in various domains} that are like to embrace LLMs as agents, and \textit{complicated and practically occurring format-oriented  instructions} so that the LLMs can be tested with real-world complex context. 
To achieve these, we implement an AI-Human collaborative strategy for developing \dataname~, featuring a structured hierarchical layout through a three-step process: i) identification and collection of domains and subdomains that AI Agents can assist humans; ii) creation of data formats specific to each subdomain; iii) generation of format-oriented instructions (\formatinstru) that include complex format requirements and real-world context for each (domain, subdomain, data format).

In experiments, we test a diverse set of closed-source LLMs (e.g., GPT-4~\citep{openai2023gpt4}, PALM2~\citep{anil2023palm}, Gemini~\citep{google2023gemini}) and open-source LLMs, such as Llama 2~\citep{touvron2023llama}, WizardLM~\citep{xu2023wizardlm}, Mistral~\citep{jiang2023mistral},  etc., on these \formatinstru~. We use GPT-4 as the main judge and human evaluation is conducted to ensure a high GPT4-Human agreement. 
The following three  observations are highlighted: 
\begin{itemize}

\item For most of the LLMs tested, the rankings of their performance on format-following do not consistently align with their rankings on content-following benchmarks (e.g., AlpacaEval). In other words, LLMs achieving good content-following performance might perform poorly on format-following and vice versa.

\item Regarding format-following, open-source LLMs lag notably behind closed-source models like GPT-3.5 and Google's Gemini, in spite of the fact that  they all exhibit similar performances on content-following benchmarks.

\item The format-following capability of LLMs varies widely across domains. Even LLMs with similar overall benchmark accuracy may demonstrate significant differences in accuracy within specific domains.

\end{itemize}

These observations suggest two key points: i) The format-following capacity of LLMs appears independent of other capabilities and may necessitate specialized alignment fine-tuning beyond the conventional instruction-tuning used in training open-source LLMs. ii) Format-following capacity is not universally transferable across domains, highlighting the potential utility of our benchmark as a guiding and probing tool for selecting domain-specific AI agent foundation models.

To our knowledge, this is the first work that breaks down LLMs' instruction-following behavior into content-following and format-following, and benchmarks the evaluation of format-following capacity. This study contributes valuable insights into comprehending LLMs' capabilities and offer guidance in selecting LLMs, particularly for domain-specific agent development. Our \dataname~dataset will be public released.

\section{Related Work}

\textbf{\contentfollowing~  benchmarks.} There are various existing efforts that build evaluation data sets to try to assess LLMs' general problem-solving capability through conversations. 
Among them, MMLU dataset~\citep{hendrycks2020measuring} is collected to measure knowledge and problem solving capabilities of LLMs in different knowledge domains such as elementary mathematics and US history. The performance of LLMs on MMLU is measured by the accuracy of their selected answers from multiple options.
AlpacaEval~\citep{dubois2023alpacafarm} and MT-Bench~\citep{zheng2023judging}, on the other hand, collect open-ended questions accross different domains without providing concrete reference answers. They rely on LLMs such as GPT-4 to conduct automatic evaluations on the target LLM's answers. 
Except for these general benchmarks, there are also evaluation data sets specifically focusing on assessing LLM answers' truthfulness~\citep{si2023large,lin2022truthfulqa} and safety~\citep{bhardwaj2023red, zhang2023safetybench, zheng2023trafficsafetygpt}. 

\textbf{\formatfollowing~ benchmarks.} 
In the past few months, several instruction-following benchmarks are recently curated to contain a small sub-set of test cases relevant to format following~\citep{zhou2023instruction,chen2024benchmarking}, such as generating data following JSON format, or following format requirements such as numbers of bullet points/paragraphs~\citep{zhou2023instruction}.
Compared to such format-following sub-sets that only covers a handful of generic formats such as JSON, our benchmark covers more diverse and domain-specific format requirements and each test example in our benchmark comes with complicated combined requirements and domain-specific context.
Therefore, we empirically find that our format-following benchmark is harder for existing LLMs and can unveil performance discrepancy across different domains,  compared to the format-following sub sets in these existing benchmarks.

\begin{table*}[ht]
\centering
\resizebox{0.9\linewidth}{!}{\small
\begin{tabular}{>{\raggedright}p{0.25\linewidth} >{\raggedright\arraybackslash}p{0.75\linewidth}}
\toprule
Domains& Subdomains\\
\midrule
\multirow{2}{*}{Healthcare} & Medical Diagnostics; Medical Treatment; Patient Care Management; Clinical Trial Analysis; Pharmaceuticals \\
\midrule
\multirow{2}{*}{Finance} & Fraud Detection; Algorithm Trading; Personalized Financial Advice; Risk management; Regulatory Compliance \\
\midrule
Technology and Software & Web Design; Programming; UI/UX Design; Data Analysis; Testing \\
\midrule
\multirow{3}{*}{Commerce and Manufacturing} & E-commerce Personalization; Manufacturing Process Optimization; Inventory and Supply Chain Management; Quality Control; Smart Logistics and Route Optimization \\
\midrule
Customer Relationship Management (CRM) & Customer Service; Sales Forecasting; Recruitment Assistants; Project Management; Lead Scoring \\
\midrule
\multirow{2}{*}{Marketing} & Consumer Behavior Analysis; Advertising Campaign Optimization; Content Curation and Creation; Social Media Management; Search Engine Optimization \\
\midrule
Scientific Research and Development & Mathematical Research; Physics; Chemistry and Biological Sciences; Environmental Sciences and Climate Change; Space Exploration \\
\midrule
\multirow{2}{*}{Education} & Adaptive Learning Platforms; Intelligent Tutoring Systems; Automated Grading Systems; Education Data Analysis; Language Learning \\
\midrule
\multirow{2}{*}{Legal} & Contract Review and Analysis; Legal Document Automation; Legal Research; Predictive Legal Analytics; Intellectual Property (IP) Management \\
\midrule
\multirow{2}{*}{Arts and Entertainment} & Music Composition; Film Scriptwriting; Visual Art Creation; Video Game Development; Sports Analytics and Performance \\
\bottomrule
\end{tabular}
}
\vspace{-.5em}
\caption{Full list of domains and subdomains.}
\label{tab:domain_list}
\end{table*}

\section{Dataset Curation and Evaluation}
\subsection{\dataname~Construction}
\label{sec:data_create}

The construction of \dataname~unfold in three steps: i) collecting domains \& subdomains, ii) gathering domain-specific data formats (each format is expressed by a name), and iii) generating \formatinstru~for each (domain, subdomain, format) triplet. Next, we elaborate on each step in detail.

\paragraph{Collecting domains \& subdomains.} To collect the domain and subdomain list for our benchmark, we adopt an iterative methodology that synergizes human expertise with the advanced capabilities of GPT-4. This process starts with an initial list of subdomains identified by domain experts. These subdomains, including but not limited to ``Web Design'', ``Programming'', and ``Medical Diagnostics'', represent areas where LLM agents have shown significant potential. This preliminary list acted as a foundation for the next phase, in which we steer GPT-4 to extend this collection by the following prompt:
\fbox{%
 \begin{minipage}{0.95\linewidth}
\textit{Can you list the domains that AI agents might help? These are some examples: Web Design, Programming, and Medical Diagnostics. Expand beyond these fields to cover all potential domains.}
 \end{minipage}
}

Following the expansion of subdomains, we utilize GPT-4 to summarize these subdomains with an instruction like: ``Summarize these domains into super domains''. Subsequently, human experts conduct a thorough review of these proposed domains and subdomains. Any domain or subdomain misaligned with our benchmark's objectives triggers a reiteration of GPT-4 to regenerate and fill in a proper domain/subdomain, followed by a subsequent expert review. This iterative cycle of generation, review, and refinement, bridging human intellectual finesse with AI efficiency, results in a well-defined list of domains and subdomains. In the end, we obtain 10 domains with each having 5 subdomains, as listed in Table \ref{tab:domain_list}.


   
                    

\paragraph{Gathering domain-specific data formats.} 
For each identified domain and subdomain, we ask GPT-4 to generate five human-understandable text data formats that an AI agent is likely to encounter. We restrict the output data formats to text-only, ensuring they are producible by LLMs. Additionally, we instruct GPT-4 to skip generic formats in this generation process to prevent the production of similar data formats across different subdomains, such as TXT, CSV, and XML. The concrete prompt is:
\fbox{%
 \begin{minipage}{0.95\linewidth}
\textit{Please give 5 human-understandable text data formats that an AI agent in the domain of \{\colorbox{blue!25}{domain}\} -> \{\colorbox{purple!25}{subdomain}\} would likely encounter as its required output formats during its interaction with humans. Note that only text data format should be provided. The data formats should also be as domain-specific as possible. Generic formats such as TXT, CSV, JSON, XML, etc, shouldn't be included. An example of a piece of data of a specific format should be provided after the name of each format.}
 \end{minipage}
}
After generating these data formats, we engage human experts to assess the quality of each format. We either regenerate or remove the data format if it does not align with the requirements described above. Some examples of domain-specific data formats are listed in Table \ref{tab:data_format}. For the full list, please check the Appendix \ref{app:data_formats}. 
From Table \ref{tab:data_format} we can see that among the generated domain-specific formats, some are existing formats that have relatively fixed and commonly acknowledged configurations, such as MathML and Maple, etc. Others are format names that would require further detailed specifications, such as Manufacturing Reports Format and Prescription Format, etc.
For both of the two categories of generated formats, we will include enough detailed format specifications in our next step of generating test prompts.
After fixing these domain-specific formats, we also add back prevalent/universal data formats applicable across all domains/subdomains, including JSON, XML, CSV, Markdown, and YAML. 


\begin{figure}[!t]
    \centering
    \includegraphics[width=0.48\textwidth]{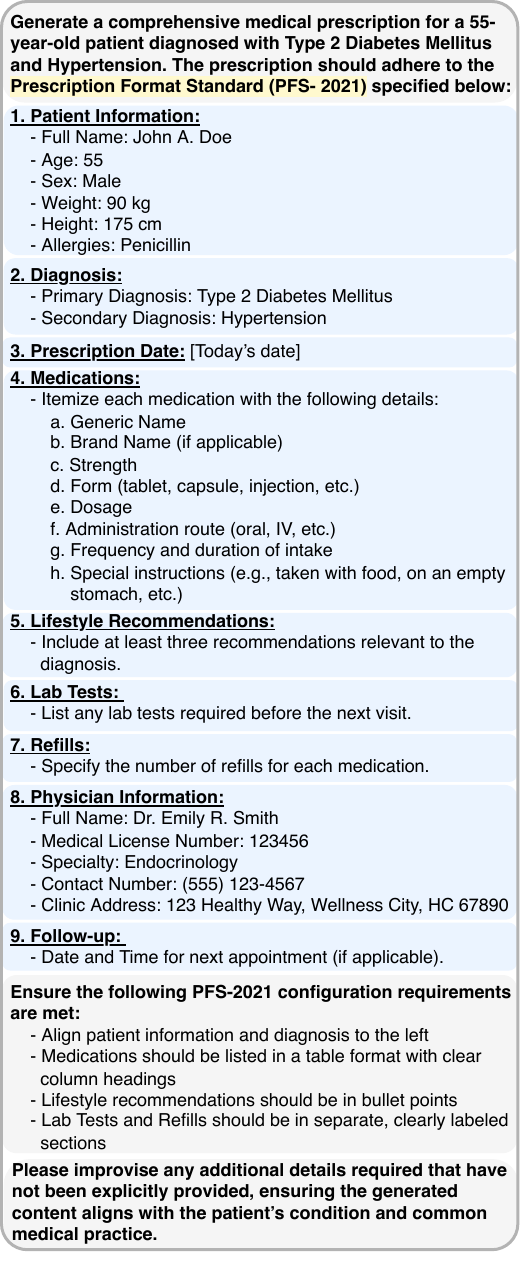}
    \caption{An \formatinstru~example when ``Domain'' = ``Healthcare'', ``Subdomain'' = ``Medical Treatment'', and ``Format'' = ``Prescription Format''. It has many detailed format requirements and missing one single format requirement would cause the target LLM to fail on this example, making our benchmark harder to ace for LLMs. }
    \label{fig:instruct}
\end{figure}
\paragraph{\formatinstru~generation.} The final phase of developing the \dataname~benchmark entails the utilization of GPT-4 to devise \formatinstru~spanning a wide array of domains, subdomains and target formats. The following prompt is employed:
\fbox{%
 \begin{minipage}{0.95\linewidth}
\textit{You are a helpful agent. Please write an instruction for an AI agent in the domain of \{\colorbox{blue!25}{domain}\} -> \{\colorbox{purple!25}{subdomain}\}. The task of the instruction should be detailed and complicated content generation in the given domain. The task should require the output to strictly adhere to a \{\colorbox{yellow!25}{format}\} format with specific configurations. If the format name is not specific enough, please give concrete illustrations of the specific format requirements to follow. Please try your best to give detailed dummy context/data required in the prompt when necessary. If you cannot give all necessary dummy data, please mention in the instruction that the AI agent is allowed to make up the data required and improvise on ungiven details. Your response should only contain the instruction or question, without any preliminary or concluding statements.}
 \end{minipage}
}

\begin{table*}[]
\centering
\resizebox{0.9\linewidth}{!}{%
\begin{tabular}{
	@{}p{0.14\textwidth}p{0.22\textwidth}p{0.35\textwidth}p{0.23\textwidth}@{}} 
	\toprule
	Domains                        & Healthcare               & Commerce and Manufacturing      & Scientific R\&D           \\ 
	\midrule
	Subdomains                    & Medical Treatment        & Manufacturing Process Optimization                  & Mathematical Research                                       \\
	\midrule
	\multirow{5}{*}{Data Formats} & Medical Reports          & Manufacturing Reports Format      & LaTeX    \\
	& Prescription Formats     & Bill of Materials   & MathML   \\
	& SOAP Notes               & Work Instruction Format        & SageMath Notebooks  \\
	& Discharge Summary        & Standard Operating Procedure & Maple     \\
	& Clinical Trial Protocols & Production Scheduling Format & MATLAB scripts   \\
	\bottomrule
\end{tabular}
}
\vspace{-.5em}
\caption{Examples of domain specific data formats under different domains and subdomains. Among the generated domain-specific formats, some  have relatively fixed and commonly acknowledged configurations, such as MathML and Maple, etc. Others are format names that would require further detailed specifications, such as Manufacturing Reports Format and Prescription Format, etc.
For both of the two categories of generated formats, we generate enough detailed format specifications in our next step of generating test prompts.}
\label{tab:data_format}
\end{table*}

These \formatinstru~are designed with a strict requirement for the outputs to conform precisely to specified data formats. In instances where the name of the data format lacks specificity, we direct GPT-4 to provide detailed examples illustrating the exact format specifications that must be adhered to. We request GPT-4 to construct tasks of considerable complexity, ensuring that the generated \formatinstru~are sufficiently detailed. Furthermore, we instruct GPT-4 to include comprehensive dummy context or data, deemed essential for the completion of these tasks. This approach not only fosters a rigorous evaluation framework for LMs but also simulates a diverse array of real-world scenarios, thereby enhancing the benchmark's relevance and applicability. Additionally, we engage human experts to verify each generated instruction, with the authority to edit, remove, or regenerate the instruction as necessary to maintain the quality of \formatinstru.

An illustrative \formatinstru~is depicted in Figure \ref{fig:instruct}, situated in the \emph{Healthcare} domain, \emph{Medical Treatment} subdomain, and \emph{Prescription Format}. 
The instruction requires generating a comprehensive prescription for a patient with Type 2 Diabetes Mellitus and Hypertension, challenging LLMs to adhere to intricate specifications. 
We can see from this example that it has many detailed format requirements, such as the content of each section, how to itemize generated content under each section, detailed FS-2021 configurations, what content to replace and what not to replace, etc. Missing one requirement would cause the target LLM to fail on this example. 
All of our generated \formatinstru~ are of similar complexity level. More examples are illustrated in Appendix~\ref{app:prompts}.

After these three steps, our \dataname~dataset is finalized, with detailed statistics in Table \ref{tab:statistics}. The average length of {\formatinstru} is the number of characters to maintain consistency with AlpacaEval.

\subsection{Evaluation Metric}
\label{sec:eval_method}
Given each \formatinstru~and corresponding LLM's response, we model the evaluation as a binary classification problem with the further requirement of generating a detailed explanation for its assessment.
GPT-4 is used as the evaluator, akin to methodologies employed in AlpacaEval and MTBench, to minimize annotation efforts. The structure of the evaluation prompt is detailed in Figure~\ref{fig:eval_temp} in Appendix, where we outline how GPT-4 is directed to assess the fidelity of responses from different LLMs to predefined format requirements. Missing a single specific format requirement among all the requirements in a \formatinstru~ would lead to failing on this prompt, making our benchmark harder to ace for LLMs. 
Our benchmark specifically focuses on assessing the ability of LLMs to comply with given format guidelines, underscoring the importance of format over content.

\section{Experiments}


\begin{table}[]
\centering
\resizebox{0.9\linewidth}{!}{
\begin{tabular}{@{}lc@{}} 
\toprule
Attributes & Number \\ 
\midrule
\# Domains                        & 10     \\
\# Subdomains                     & 50     \\
\# Data Formats                   & 248    \\
\# \formatinstru                  & 494    \\
Average Length of \formatinstru & 2,908  \\
\bottomrule
\end{tabular}
}
\vspace{-.5em}
\caption{\dataname~statistics.}
\vspace{-1em}
\label{tab:statistics}
\end{table}

\begin{table*}[] 
\centering
\resizebox{0.9\linewidth}{!}{
\begin{tabular}{
  llccc
}
\toprule
\multicolumn{2}{c}{\multirow{2}{*}{Model}} & \formatfollowing~& \multicolumn{2}{c}{{\contentfollowing}} \\
\cmidrule(lr){3-3} \cmidrule(lr){4-5}
&  & {\dataname}  & {MT-Bench} & {AlpacaEval} \\
\midrule
\midrule
\multirow{9}{*}{\rotatebox{90}{\small{Open-source}}} & Vicuna 13B V1.3~\citep{chiang2023vicuna} & 22.74  & 6.39 & 82.11 \\
& WizardLM 13B V1.1~\citep{xu2023wizardlm} & 27.00 & 6.76 & 86.32 \\
& Vicuna 13B V1.5-16k~\citep{chiang2023vicuna} & 27.08  & 6.92 & - \\
& Openchat V3.2-super~\citep{wang2023openchat} & 31.22  &7.19 & 89.50 \\
& Llama 2 7B Chat~\citep{touvron2023llama} & 45.44  &6.27 &71.37  \\
& Mistral 7B Instruct V0.1~\citep{jiang2023mistral} & 46.91  &6.84 &92.78 \\
& Llama 2 13B Chat~\citep{touvron2023llama} & 53.28  & 6.77 &81.09 \\
& WizardLM 13B V1.2~\citep{xu2023wizardlm} & 63.54  &7.2 &89.17 \\
& Zephyr 7B Beta~\citep{tunstall2023zephyr} & 64.12  &7.34 &90.60  \\
\midrule
\midrule
\multirow{4}{*}{\rotatebox{90}{\small{Closed-source}}} & GPT-3.5~\citep{openai2023chatgpt} & 80.66  & 8.32 & 93.42 \\
& Gemini Pro~\citep{google2023gemini} & 80.25  & - & 79.66 \\
& PaLM 2 for Text 32k~\citep{anil2023palm}  & 83.72 & - &   -   \\
& GPT-4~\citep{openai2023gpt4} & 91.17  & 9.18 & 95.28 \\
\bottomrule
\end{tabular}
}
\vspace{-.5em}
\caption{Main Results. \jiangshu{The source of the models can be found in Appendix~\ref{app:eval_models}.}}
\vspace{-.5em}
\label{tab:main_result}
\end{table*}

In this section, we present our empirical results and analysis of {\dataname}. 
To verify {\dataname}'s effectiveness on serving as a format-following benchmark of LLMs, we firstly select top-performing LLMs from both the closed-source and open-source world and evaluate them on {\dataname}.
Specifically, we are more interested in middle-sized LLMs that have shown similar performances to GPT-3.5 and GPT-4 on existing content-following benchmarks such as MT-Bench and AlpacaEval because such LLMs are currently most widely-used.

To evaluate each open-source LLM, we employ its official prompt format to conduct generation given each test prompt in {\dataname}. During generation, we use sampling and set the temperature as 0.7 for all models for fair comparison. We also set the max new tokens to generate as 5120 for all models. After generation, we evaluate each LLM's format-following accuracy with GPT-4 as judge, as illustrated in Section~\ref{sec:eval_method}.

\begin{figure*}[t]
\centering
    \resizebox{0.9\linewidth}{!}{
     \begin{tabular}{cc}
     \includegraphics[width=.33\linewidth]{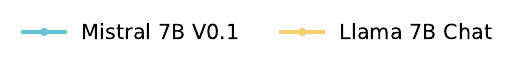}& 
     \includegraphics[width=.37\linewidth]{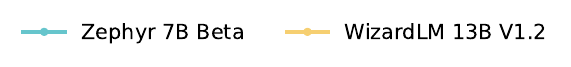} 
     \vspace{-.8em}\\
     \includegraphics[width=.45\linewidth]{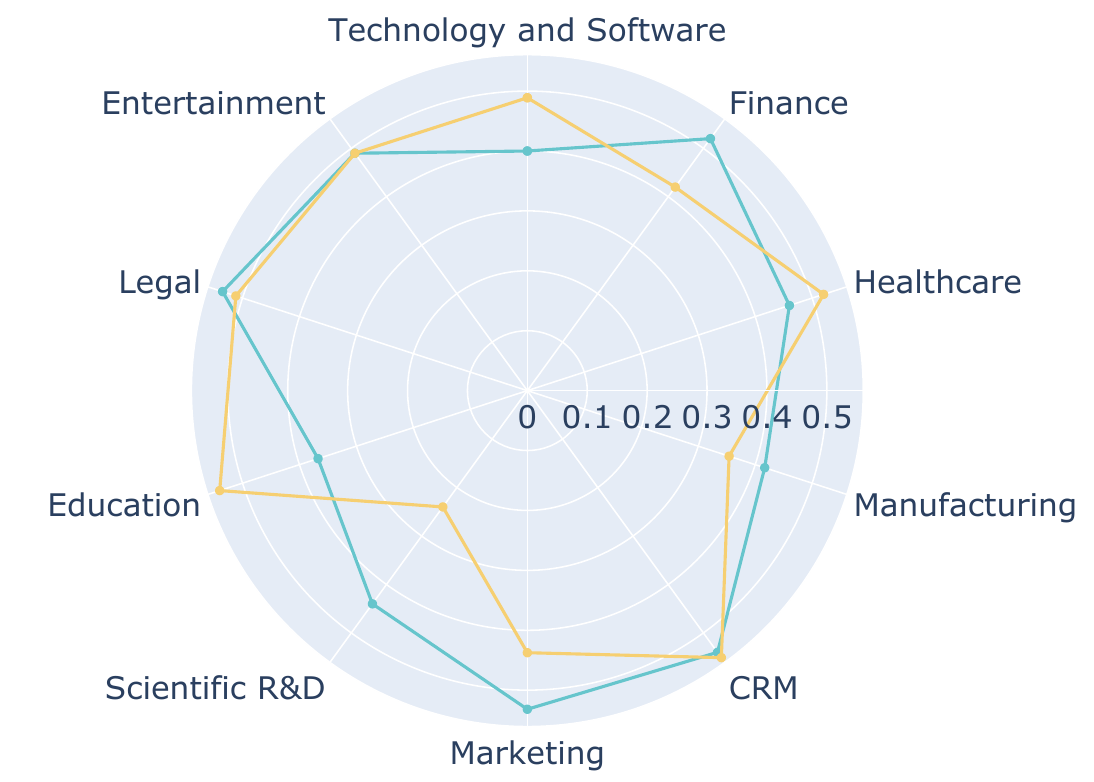}& 
     \includegraphics[width=.45\linewidth]{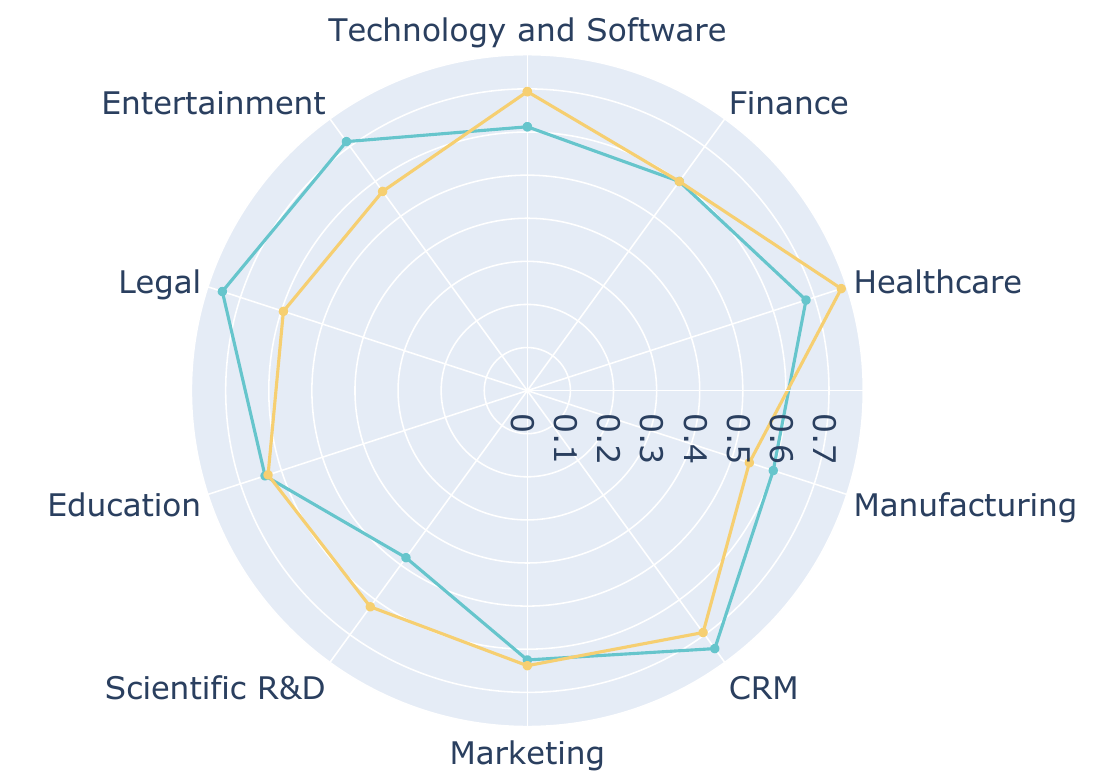}\\
     \footnotesize{(a) Mistral 7B V0.1 VS Llama 7B Chat} & \footnotesize{(b) Zephyr 7B Beta VS WizardLM 13B V1.2}
     \end{tabular}
     }
     \vspace{-.5em}
     \caption{Domain Analysis on different models with similar performance.}
     \vspace{-.5em}
     \label{fig:domain_analysis}
\end{figure*}

\begin{figure*}[t]
\centering
    \resizebox{0.9\linewidth}{!}{
     \begin{tabular}{cc}
     \includegraphics[width=.35\linewidth]{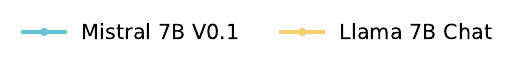} & 
     \includegraphics[width=.27\linewidth]{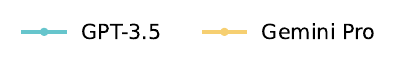}
     \vspace{-.9em}\\
     \includegraphics[width=.44\linewidth]{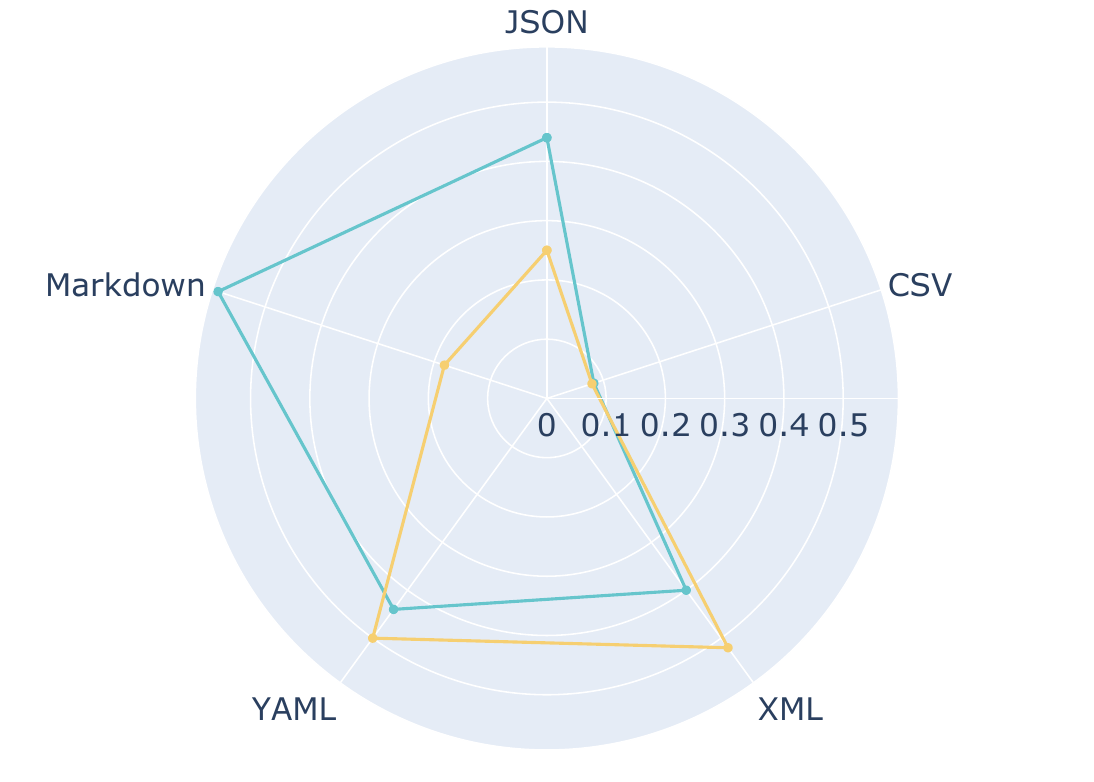}& 
     \includegraphics[width=.44\linewidth]{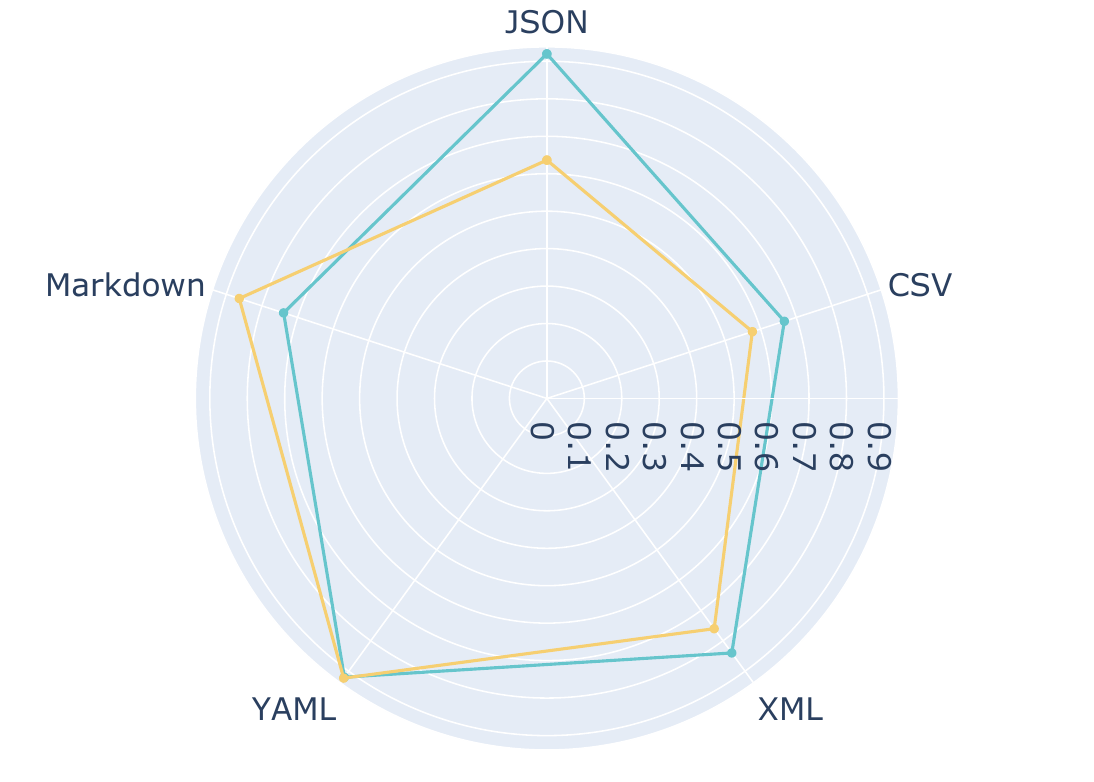}\\
     \footnotesize{(a) Mistral 7B V0.1 VS Llama 7B Chat} & \footnotesize{(b) GPT-3.5 VS Gemini Pro}
     \end{tabular}
     }
     \vspace{-.5em}
     \caption{Data format Analysis on different models with similar performance.}
     \vspace{-.5em}
     \label{fig:format_analysis}
\end{figure*}



\subsection{Main Results}
Table~\ref{tab:main_result} presents the format accuracy of all selected LLMs on {\dataname}. To better see the performance discrepancy of the same LLM on different benchmarks, we also list each LLM's performance on AlpacaEval and MT-Bench in Table~\ref{tab:main_result}.

The first observation we have from Table~\ref{tab:main_result} is that performance rankings of LLMs on {\dataname} is not consistent with their rankings on content-following evaluation benchmarks. 
For example, Openchat V3.2-super and WizardLM 13B V1.2 have similar performance on both AlpacaEval (around 89\%) and MT-Bench (around 7.2). While their format accuracy on {\dataname} has more than 30 points gap (31.22\% vs 63.54\%). Gemini-pro has lower performance compared to GPT-3.5 on AlpacaEval while their format accuracy on {\dataname} is similar. We can see similar patterns when viewing {\dataname} results side-by-side with those on AlpacaEval 2.0.

Second, we have found that closed-source models such as GPT-3.5 and Gemini Pro, significantly outperform open-source models. Closed-source models all have above 80\% of format accuracy while open-source models reach only below 70\% of format accuracy. This performance gap is much more significant compared to their performance gap on existing content-following benchmarks such as AlpacaEval.
For example, Mistral 7B Instruct V0.1 achieves 92.78\% of accuracy on AlpacaEval, comparable with GPT-3.5's 93.42\%. While it's format accuracy on {\dataname} is 46.91\%,  much lower compared to GPT-3.5's 80.66\%. Other open-source models also show similar performance pattern.

These two observations suggest that format-following capability is independent of other capabilities of LLMs reviewed by existing content-following evaluation benchmarks. It  might require tailored alignment fine-tuning with specific data beyond regular instruction-tuning that are widely used for fine-tuning open-sourced LLMs.


\subsection{Result Analysis}
\noindent\textbf{Domain Analysis.} When analyzing empirical results on {\dataname}, we have also noticed that the format-following capability of LLMs may vary a lot across different domains. For two LLMs that give similar final accuracy on {\dataname}, their accuracy on different domains can be very different.  
Figure~\ref{fig:domain_analysis} shows two examples. Figure~\ref{fig:domain_analysis}(a) shows the format-following accuracy comparison between Mistral 7B V0.1 and Llama 7B Chat across different domains. Their final performance on {\dataname} are very similar and both around 46\%. While we can see from  Figure~\ref{fig:domain_analysis}(a) that Llama 7B Chat performs significantly worse on Scientific Research and Development domain and performs much better on Education domain. 
Similarly, Zephyr 7B Beta and WizardLM 13B V1.2 have similar final performance on {\dataname} (around 64\%) while Figure~\ref{fig:domain_analysis}(b) shows they have their own expertise on different domains.
It indicates that format-following capacity is not generalizable across domains, possibly due to domain-specific formats, such as SOP format in Commerce and Manufacturing, MathML and Maple format in Scientific R\&D, etc. It shows that our benchmark can potentially serve as guidance and probing tools for the choice of domain-specific AI agent foundation models.

\noindent\textbf{Format Analysis.}  As mentioned in Section 3, except domain-specific formats, {\dataname} also includes 5 general formats (JSON, XML, CSV, Markdown, and YAML.) and create {\formatinstru} with general formats and domain-specific context. 
After seeing the performance discrepancy across domains, we are also curious about whether LLMs have their own expertise on different general formats. Therefore in Figure~\ref{fig:format_analysis}, we present the performance comparison of two LLMs achieving similar final accuracy across different general formats. 
Figure~\ref{fig:format_analysis}(a) shows that both Mistral 7B V0.1 and Llama 7B chat don't perform well on managing CSV format. While Mistral 7B V0.1 is good at JSON and Markdown and Llama 7B chat expertizes in YAML and XML.
Figure~\ref{fig:format_analysis}(b) shows that both GPT-3.5 and Gemini Pro performs well on following YAML format and GPT-3.5 is more specialized on JSON. 
It indicates that different models have their own expertise on formats as well. If multiple general formats are suitable for one target generation task, our benchmark can be used to pick the best format for a LLM to reduce errors on format-following on the target task.

\noindent\textbf{Error Analysis.} Table \ref{tab:main_result} reveals that many open-source LLMs underperform on our Format Following ({\dataname}) benchmark. To understand these shortcomings, we performed an error analysis focusing on Mistral 7B Instruct V0.1 as a representative model. We request human annotators to examine the explanations given by GPT-4 for the failure instances and categorize these reasons into the following groupings: 1) \textit{Incomplete Sections}: The model often neglects essential sections that the prompts mandate, spanning a variety of content areas such as methodological frameworks, data analyses, theoretical discussions. This shortfall is especially pronounced in scientific contexts, as illustrated in Figure~\ref{fig:domain_analysis}(a), which depicts Mistral 7B V0.1's markedly inferior performance in scientific domains. 2) \textit{Incorrect Data Structure}: Mistral 7B V0.1 struggles with adhering to specific structural guidelines, impacting diverse formats including JSON, CSV, Markdown (as highlighted in Figure~\ref{fig:format_analysis}(a)), as well as domain-specific formats such as legal citations and academic referencing. Issues include improper syntax use, inaccurate data structure representations, and failure to follow document layout guidelines. 
3) \textit{Missing Detailed Format Requirements}: The model frequently fails to meet specific and detailed formatting requirements. For instances, it presents the "Age" attribute in a text format ("40 years") rather than as a numerical value (40); it ignores the directive to use bullet points ("-") for lists and enumerations; it incorrectly includes headers, despite explicit instructions in the prompt to exclude them.

\noindent\textbf{Human Evaluation Alignment.} To evaluate the effectiveness of utilizing GPT-4 as evaluator, we randomly selected 100 annotations made by GPT-4 on the outputs of Mistral 7B V0.1 for a comprehensive human appraisal. We engaged five human experts to review the accuracy of GPT-4's annotations and conducted an analysis to compare the agreement between human annotators and the GPT-4 evaluator.
Our findings revealed that for 84 out of the 100 {\formatinstru}, the evaluations by GPT-4 were in agreement with those of the human experts, yielding an alignment rate of 84\%. A closer examination was conducted on the 16 instances where GPT-4's evaluations diverged from human judgment. It was observed that all these discrepancies were instances of false positives, indicating scenarios where the model's outputs failed to adhere to specific format requirements, yet were overlooked by GPT-4. For example, one {\formatinstru} required the inclusion of an address, which was absent in the model's output.
Additional errors included the generation of an insufficient number of examples, omission of a required section, and the introduction of non-existent tags. Consequently, these findings suggest that the actual performance of the models might be lower than what is reported by GPT-4's evaluations. 

Taking human evaluations as a benchmark, we infer that the real performance of the models could be approximately 16\% lower than the figures presented in Table~\ref{tab:main_result}. For example, the accuracy of the Mistral 7B Instruct V0.1 model, as evaluated by GPT-4, stands at 46.91\%, but its actual performance is estimated to be around 39.4\%. While employing GPT-4 as an evaluator has its limitations, it significantly reduces the workload associated with human evaluation and provides insights into the comparative performance of different models. For instance, our benchmark indicates that Zephyr 7B Beta is the best open-sourced LLMs in Table \ref{tab:main_result}.


\noindent\textbf{Comparison with IfEval.} When generating the {\formatinstru} of {\dataname}, we prompt GPT-4 to create instructions that contain detailed and complex content generation tasks with specific format configurations, as shown in Figure~\ref{fig:instruct}.
This is one of the main differences that {\dataname} has compared to other  benchmarks such as IfEval. Although IfEval has a sub-set of prompts that test the format-following capability of LLMs, they are domain-agnostic and contain relatively simple rule-based format configurations.
We evaluate representative open-source and closed-source LLMs on IfEval's detectable format sub set too and find that they achieve much higher accuracy on IfEval's sub-set compared to on {\dataname}, as shown in Table~\ref{tab:ifeval_result} in Appendix. 
It indicates that {\dataname} is not only the first domain-specific format-following benchmark, but also a much harder one compared to current format-following test sets.


\noindent\textbf{Cost Analysis.} In this work, we utilize GPT-4 API for both the creation of our benchmark and the evaluation of LLMs, thereby incurring associated expenses. We estimate the cost of generating the {\dataname} benchmark to be approximately \$25. Furthermore, the expense of evaluating a single LLM on {\dataname} is estimated to be around \$40. In future endeavors, we plan to consider using GPT-4-Turbo as the evaluator to reduce costs. Please refer to Appendix~\ref{app:cost} for more details.

\section{Conclusion}
In conclusion, our introduction of {\dataname} marks a significant advancement in evaluating large language models' (LLMs) format-following capabilities, a crucial but previously overlooked aspect. Through a novel AI-Human collaborative construction, {\dataname} offers a comprehensive benchmark covering a diverse range of formats and instructions. Our findings reveal that format-following is an independent skill set not correlated with content generation performance, highlight a gap between open and closed-source LLMs in format adherence, and underscore the variability of LLMs' format-following proficiency across domains.

\section*{Limitations}
Our research marks a significant step forward in assessing the format-following capabilities of large language models (LLMs). However, it is not without its challenges. One constraint is our reliance on human experts for benchmark validation, including test prompt verification. This dependence could both introduce subjectivity and limit the ability to scale. To address this, future work will aim to reduce human involvement by crafting a more automated, yet equally robust, system for test case generation and validation, thereby broadening the benchmark’s applicability. Moreover, the evaluation of LLMs, particularly through the use of GPT-4 APIs, incurs some costs. An alternative we intend to explore is employing GPT-4 Turbo as a more cost-effective solution without compromising the assessment's quality. Furthermore, our benchmark, although extensive, might not fully represent the diverse array of format requirements seen in real-world settings. Continuous refinement and expansion of our benchmark are essential to more accurately reflect the vast spectrum of practical use cases, enhancing its utility for future LLM development and deployment.


\bibliography{anthology,custom}
\bibliographystyle{acl_natbib}
\clearpage
\appendix

\section{Appendix A}
\label{sec:appendix}
\subsection{Evaluated Models}
\label{app:eval_models}
We evaluate the following models with our FoFo benchmark:
\subsubsection*{Open-source LLMs}
\begin{itemize}
    \item Vicuna 13B V1.3~\citep{chiang2023vicuna}\footnote{\url{https://huggingface.co/lmsys/vicuna-13b-v1.3}}
    \item Vicuna 13B V1.5-16k~\citep{chiang2023vicuna}\footnote{\url{https://huggingface.co/lmsys/vicuna-13b-v1.5-16k}}
    \item WizardLM 13B V1.1~\citep{xu2023wizardlm}\footnote{\url{https://huggingface.co/WizardLM/WizardLM-13B-V1.1}}
    \item WizardLM 13B V1.2~\citep{xu2023wizardlm}\footnote{\url{https://huggingface.co/WizardLM/WizardLM-13B-V1.2}}
    \item Openchat V3.2-super~\citep{wang2023openchat}\footnote{\url{https://huggingface.co/openchat/openchat_v3.2_super}}
    \item Llama 2 7B Chat~\citep{touvron2023llama}\footnote{\url{https://huggingface.co/meta-llama/Llama-2-7b-chat-hf}}
    \item Llama 2 13B Chat~\citep{touvron2023llama}\footnote{\url{https://huggingface.co/meta-llama/Llama-2-13b-chat-hf}}
    \item Mistral 7B Instruct V0.1~\citep{jiang2023mistral}\footnote{\url{https://huggingface.co/mistralai/Mistral-7B-Instruct-v0.1}}
    \item Zephyr 7B Beta~\citep{tunstall2023zephyr}\footnote{\url{https://huggingface.co/HuggingFaceH4/zephyr-7b-beta}}
\end{itemize}

\subsubsection*{Closed-source LLMs}
\begin{itemize}
    \item GPT-3.5~\citep{openai2023chatgpt}\footnote{We use ``gpt-4' \url{https://platform.openai.com/docs/models/gpt-4-and-gpt-4-turbo}}
    \item GPT-4~\citep{openai2023gpt4}\footnote{We use ``gpt-3.5-turbo-1106''. \url{https://platform.openai.com/docs/models/gpt-3-5-turbo}}
    \item Gemini-Pro~\citep{google2023gemini}\footnote{\url{https://cloud.google.com/vertex-ai/docs/generative-ai/model-reference/gemini}}
    \item PaLM 2 for Text 32k ~\citep{anil2023palm}\footnote{We use ``text-bison-32k''.~\url{https://cloud.google.com/vertex-ai/docs/generative-ai/model-reference/text}}
\end{itemize}

\subsection{Cost Analysis}
\label{app:cost}
In this study, we utilize GPT-4 for both the construction of our benchmark and the evaluation of LLMs. This approach incurs certain costs associated with utilizing the GPT-4 API. The cost is determined by the number of tokens in both the input and output, with pricing set at \$0.03 per 1,000 input tokens and \$0.06 per 1,000 output tokens\footnote{\url{https://openai.com/pricing}}. Given that one token approximately equates to four characters of common English text according to OpenAI's guidelines\footnote{\url{https://platform.openai.com/tokenizer}}, we can estimate the average number of tokens based on the character count.


In the process of developing our benchmark, the primary expenditure arises during the creation of FoFo, as detailed in Section~\ref{sec:data_create}. Analysis reveals that the mean length of input characters is 818, translating to approximately 205 GPT-4 tokens. Conversely, the output character count averages at 2908, equivalent to about 727 GPT-4 tokens. This setup yields an average cost per prompt of \$0.03 * 0.205 (for the input) + \$0.06 * 0.727 (for the output), amounting to \$0.05. Considering the necessity to craft 500 such prompts, the aggregate cost allocated for the generation of test prompts stands at \$25.

In our evaluation of various LLMs using GPT-4, the inputs consist of three components as detailed in Figure \ref{fig:eval_temp}: the length of the evaluation template (which contains 1,594 characters), the length of \formatinstru~, and the length of the model outputs.
As indicated in Table~\ref{tab:statistics}, the average character length of \formatinstru~ is 2,908. The average character length of models' outputs is 4,163. Consequently, the average input character length for evaluation purposes totals 8,665, equivalent to approximately 2,166 tokens for GPT-4. This results in an average input cost of: \$0.03 * 2.166 = \$0.065. For the outputs, the average character length during evaluation is 1,098, translating to roughly 275 tokens for GPT-4. This leads to an average output cost of: \$0.06 * 0.275 = \$0.0165. Therefore, the average cost of evaluating one LLM using a single prompt stands at \$0.0815. The entire Format Following (FoFo) benchmark comprises around 500 prompts, culminating in a total evaluation cost of approximately \$40 for one LLM across the entire benchmark. In future endeavors, we are considering the adoption of GPT-4-Turbo for evaluation purposes to further mitigate costs.

\begin{table*}[] 
\centering
\resizebox{0.8\linewidth}{!}{
\begin{tabular}{
  llcc
}
\toprule
\multicolumn{2}{c}{\multirow{2}{*}{Model}} & FoFo & IfEval \\
&  & {Format Acc} &  {Acc on Detectable Formats} \\
\midrule
\midrule
& WizardLM 13B V1.1~\citep{xu2023wizardlm} & 27.00 & 59.24  \\
& Mistral 7B Instruct V0.1~\citep{jiang2023mistral} & 46.91 &60.51  \\
& WizardLM 13B V1.2~\citep{xu2023wizardlm} & 63.54 &69.43 \\
& Zephyr 7B Beta~\citep{tunstall2023zephyr} & 64.12 &66.24 \\
\midrule
\midrule
& GPT-3.5~\citep{openai2023chatgpt} & 80.66 & 91.72\\

\bottomrule
\end{tabular}
}
\caption{Comparison of LLMs' performance on IfEval format sub-set and on FoFo.}
\label{tab:ifeval_result}
\end{table*}

\subsection{Comparison with IfEval.} 
Table~\ref{tab:ifeval_result} illustrates the performance of representative open-source and closed-source LLMs on \dataname~ and the detectable formats subcategory in IfEval. As shown in Table~\ref{tab:ifeval_result}, these models achieve significantly higher accuracy on the IfEval subset compared to \dataname. Furthermore, the performance gap between different models on IfEval is much narrower compared to that on \dataname.

\subsection{Examples of \formatinstru}
\label{app:prompts}
\begin{figure*}[!t]
    \centering
    \includegraphics[width=0.95\linewidth]{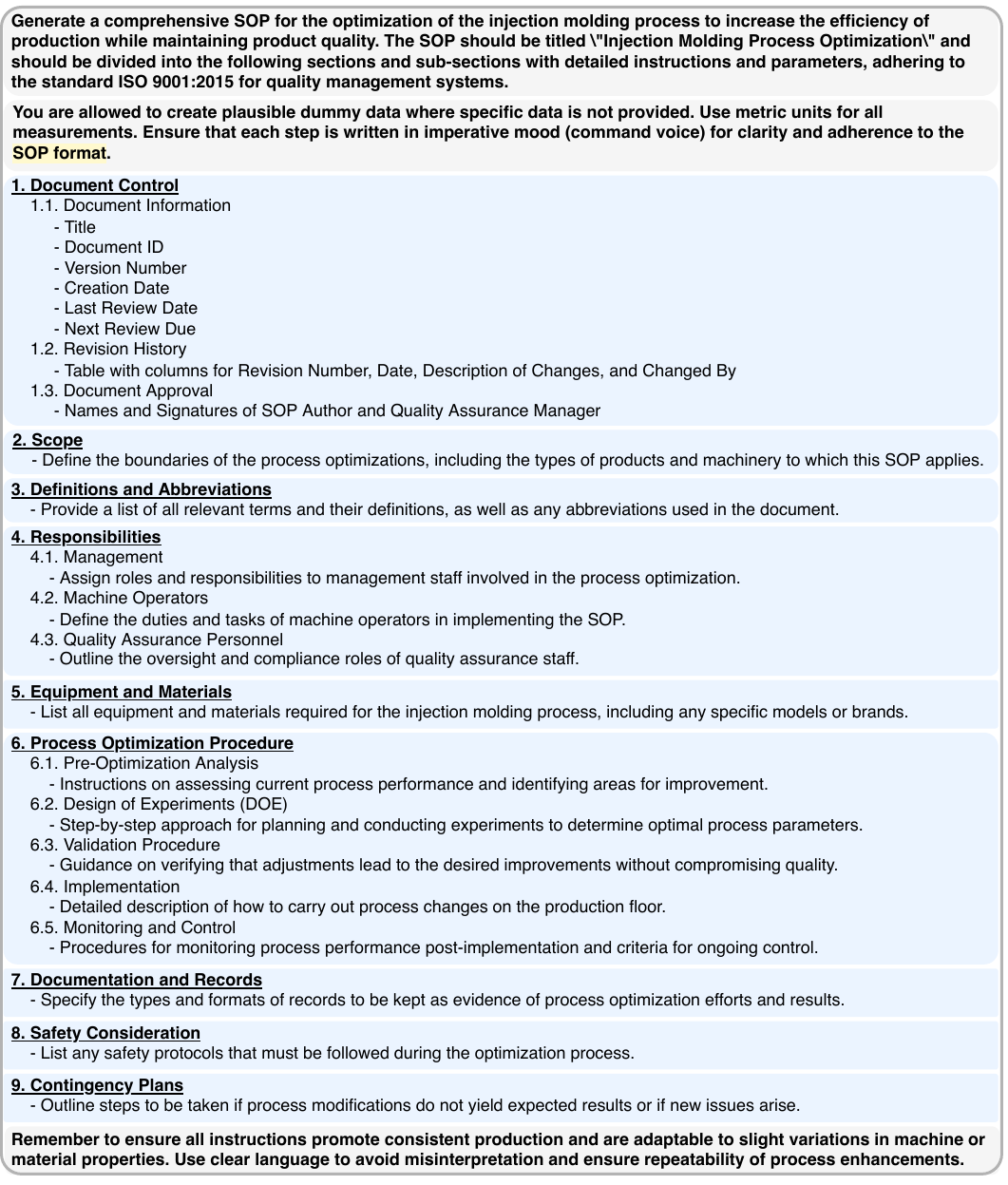}
    \caption{An \formatinstru~example when ``Domain'' = ``Commerce and Manufacturing'', ``Subdomain'' = ``Manufacturing Process Optimization'', and ``Format'' = ``Standard Operating Procedure (SOP)''.}
    \label{fig:instruct_specific_appendix}
\end{figure*}

\begin{figure*}[!t]
    \centering
    \includegraphics[width=0.95\linewidth]{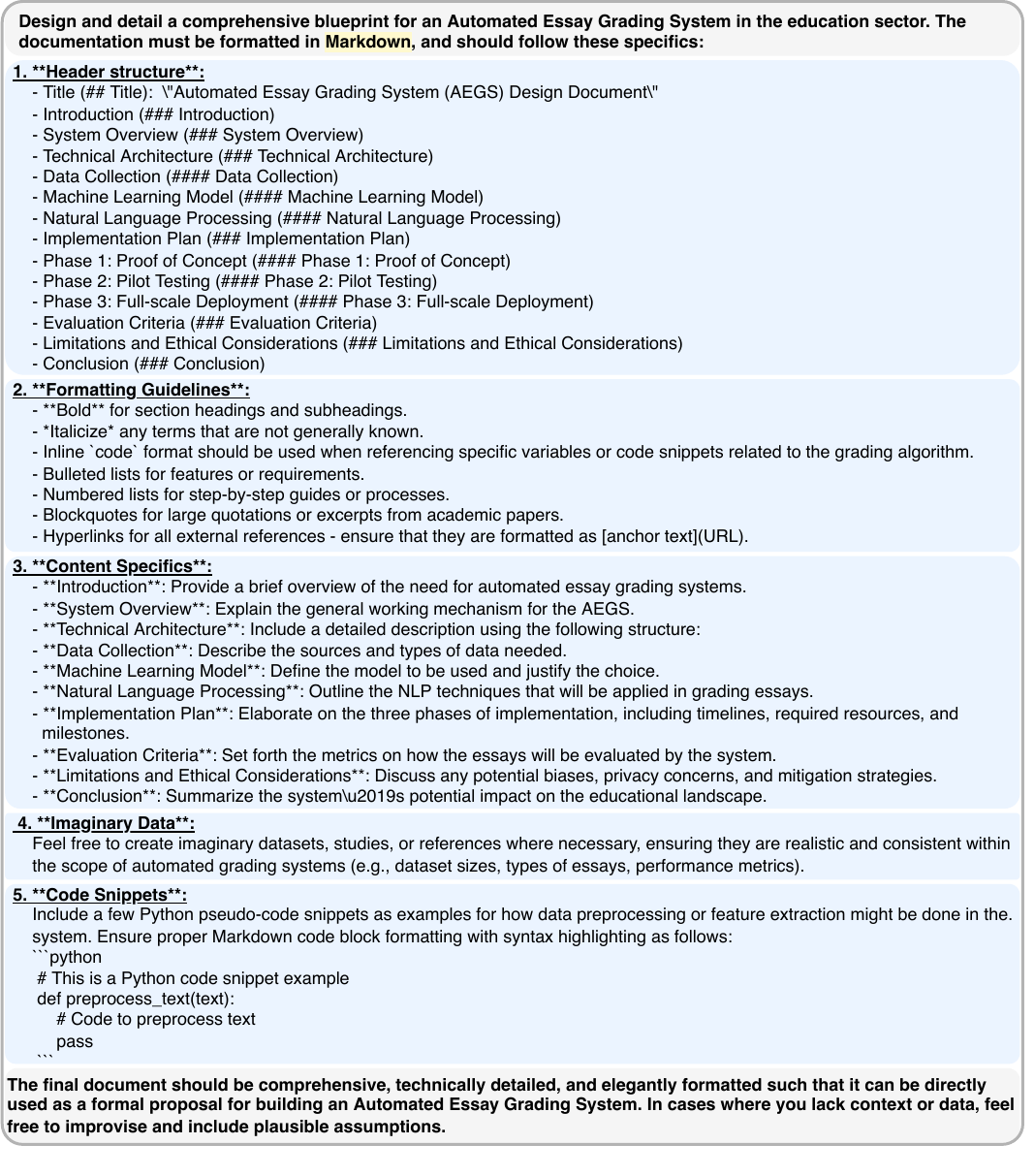}
    \caption{An \formatinstru~example when ``Domain'' = ``Education'', ``Subdomain'' = ``Automated Grading Systems'', and ``Format'' = ``Markdown''.}
    \label{fig:instruct_general_appendix}
\end{figure*}

We show more examples of \formatinstru~ in Figure~\ref{fig:instruct_specific_appendix} and Figure~\ref{fig:instruct_general_appendix}. Figure~\ref{fig:instruct_specific_appendix} shows an example when ``Domain'' = ``Commerce and Manufacturing'', ``Subdomain'' = ``Manufacturing Process Optimization'', and ``Format'' = ``Standard Operating Procedure (SOP)'', while Figure~\ref{fig:instruct_general_appendix} is an example when ``Domain'' = ``Education'', ``Subdomain'' = ``Automated Grading Systems'', and ``Format'' = ``Markdown''. Both examples include detailed enough format specifications under each domain.

\subsection{Examples of GPT-4 annotations}
\begin{figure*}[!t]
    \centering
    \includegraphics[width=\linewidth]{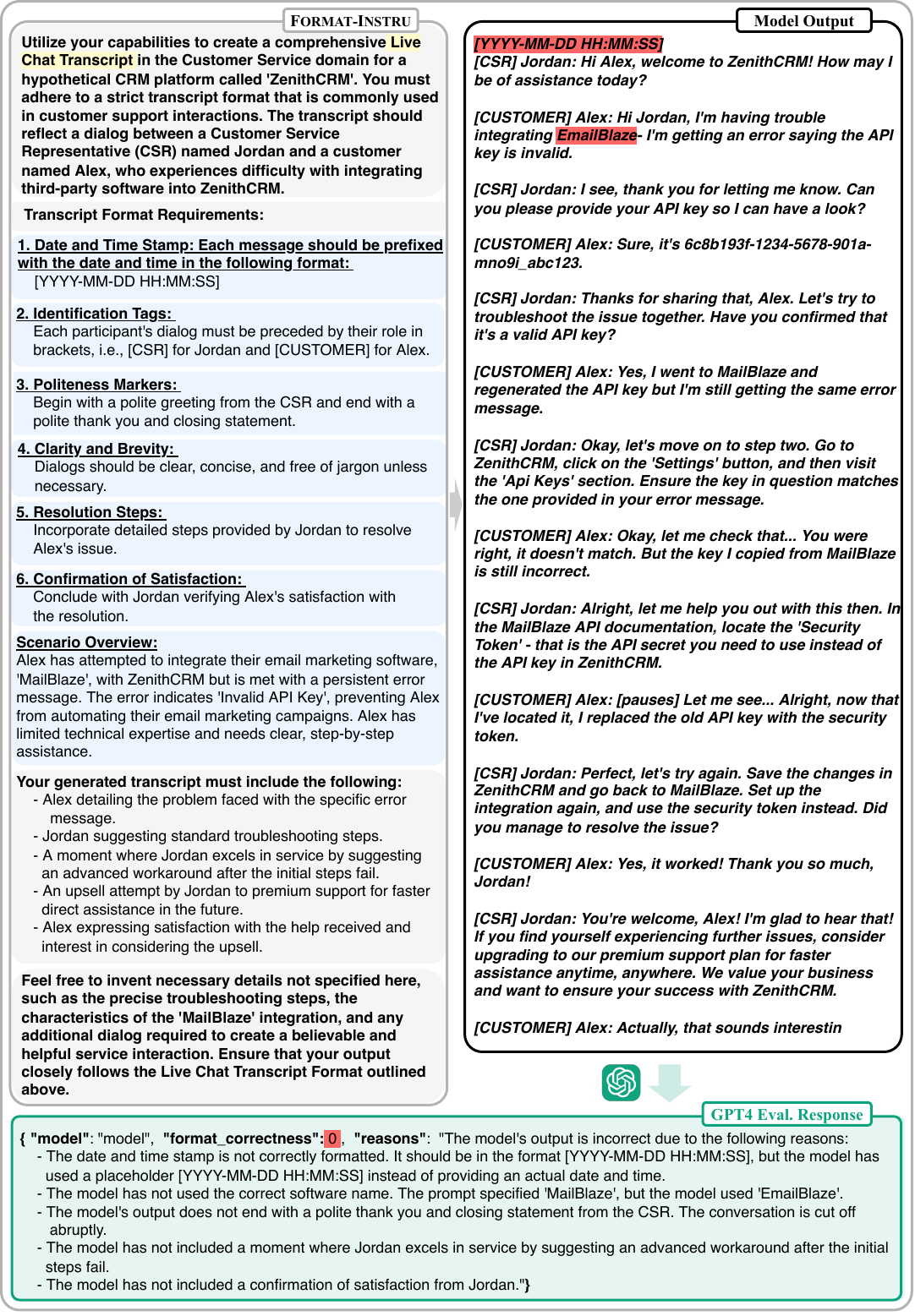}
    \caption{An example of GPT-4 annotation. In this example, the format correctness of the output from Mistral 7B Instruct V0.1, based on a given \formatinstru, is labeled as `0' (indicating `False'). GPT-4 also include the reasons why the format of the output is not correct.}
    \label{fig:annotation_example}
\end{figure*}

We also list an example of GPT-4 annotations on the output from Mistral 7B Instruct V0.1 in Figure~\ref{fig:annotation_example}. In this example, the format correctness of the output from Mistral 7B Instruct V0.1, based on a given instruction format, is labeled as `0' (indicating `False'). GPT-4 also includes the reasons why the format of the output is not correct. For example, the date and time stamp is not correctly formatted or the correct software name is not used.

\subsection{List of domain specific data formats}
\label{app:data_formats}
The full list of domain specific data formats is illustrated in page 16-18.

\subsection{Evaluation Prompt Template}
The evaluation prompt template we used to evaluate the performance of different models is shown in Figure~\ref{fig:eval_temp}.

\clearpage
\onecolumn
\begin{longtable}{>{\raggedright\arraybackslash}p{0.12\linewidth}
                  >{\raggedright\arraybackslash}p{0.18\linewidth}
                  >{\raggedright\arraybackslash}p{0.7\linewidth}}

\toprule
      Domain & Subdomain & Dataformat \\ \midrule
\midrule
\endfirsthead

\toprule
      Domain & Subdomain & Dataformat \\ \midrule
\midrule
\endhead

\midrule
\multicolumn{3}{r}{{Continued on next page}} \\ \bottomrule
\endfoot

\bottomrule
\endlastfoot

\multirow{17}{*}{Healthcare}   & {Medical Diagnostics}  & \parbox[t]{\linewidth}{
    ICD-10 format (International Classification of Diseases 10th Revision); LOINC format (Logical Observation Identifiers Names and Codes); CAP format (College of American Pathologists protocol);
    DICOM SR (Structured Report); HL7 CDA (Health Level Seven Clinical Document Architecture)
} \\
\cmidrule{2-3}
 & {Medical Treatment}     & \parbox[t]{\linewidth}{
 Medical Reports; Prescription Formats; SOAP Notes; Discharge Summary; Clinical Trial Protocols }    \\
 \cmidrule{2-3}
& Patient Care   Management       & \parbox[t]{\linewidth}{
Electronic Health Record (EHR) Format; 
Discharge Summary Format;
Clinical Trial Report Format;
Prescription Format;
Medical Coding and Billing Statement Format }  \\
\cmidrule{2-3}
 & Clinical Trial   Analysis       & \parbox[t]{\linewidth}{
 Clinical Study Report (CSR);
 Clinical Trial   Protocol;
 Patient Reported Outcomes (PRO);
 Patient Data Report (PDR);
 Adverse Event Report }\\
 \cmidrule{2-3}
& Pharmace- uticals                 & \parbox[t]{\linewidth}{
RxNorm Format; 
HL7 (Health Level 7) Format; 
ICD-10   Format; 
LOINC Format; 
Fast Healthcare Interoperability Resources (FHIR)   Format} \\ 
\midrule
\multirow{15}{*}{Finance}                   & Fraud Detection                 & \parbox[t]{\linewidth}{
Compromised Account Report Format; 
Credit Card   Fraud Alert Format; 
Investment Fraud Detection Report; 
Loan Fraud Report  Format; 
Insurance Fraud Detection Report} \\
\cmidrule{2-3}
 & Algorithm Trading               & \parbox[t]{\linewidth}{
 FIX Protocol Message; 
 Standard Portfolio Analysis of Risk Data; 
 Thomson Reuters EIKON data; 
 Bloomberg Terminal Data;   
 Morningstar Data   }\\
 \cmidrule{2-3}
& Personalized   Financial Advice & \parbox[t]{\linewidth}{
Financial Reports; 
Investment Strategy Reports;  
Personal Financial Plans; 
Risk Profile Reports; 
Asset Performance Reports}  \\
\cmidrule{2-3}
 & Risk management                 & \parbox[t]{\linewidth}{
 Financial Risk Analysis Reports; 
 Basel III   Regulatory Filings; 
 Value At Risk (VaR) Statements; 
 Credit Default Swap (CDS)   Spreads; 
 Stress Testing Reports }   \\
 \cmidrule{2-3}
& Regulatory   Compliance         & \parbox[t]{\linewidth}{
Legal Document Format; 
Financial Report Format (FRF); 
Risk Assessment Reports; 
Regulatory Filings Format; 
Compliance Audit  Reports   } \\ 
\midrule
\multirow{13}{*}{\shortstack[l]{Technology \\
and \\
Software}} & Web Design                      & \parbox[t]{\linewidth}{
HTML (HyperText Markup Language); 
CSS (Cascading   Style Sheets); 
JavaScript; 
SVG (Scalable Vector Graphics); 
.htaccess  }\\
\cmidrule{2-3}
 & Programming    & \parbox[t]{\linewidth}{
 Python Files (.py); 
 JavaScript Files (.js); 
 SQL   Files (.sql); 
 Java Source Files (.java); 
 C++ Source Code Files (.cpp)   } \\
 \cmidrule{2-3}
& UI/UX Design                    & \parbox[t]{\linewidth}{
UX/UI Mockup Annotation;
User Journey Mapping Text;   
Usability Test Session Transcripts; 
UX/UI Design Specification (Typography   Palette etc.);
User Persona Description }  \\
\cmidrule{2-3}
 & Data Analysis                   & \parbox[t]{\linewidth}{
 SQL Result Set; 
 Python Pandas DataFrame; 
 R Data   Frame Output; 
 Data Dictionary Output; 
 Log File Output  } \\
 \cmidrule{2-3}
& Testing                         & \parbox[t]{\linewidth}{
Test Plan; 
Test Case Description; 
Bug/Issue Report;   
Test Summary Report; 
Requirement Traceability Matrix (RTM)  }  \\ 
\midrule 
          & E-commerce Personalization             & \parbox[t]{\linewidth}{
E-commerce Invoice Format; 
Personalized Product Recommendation; 
Shopping Cart Abandonment Reminder; 
Shipping and Delivery Notification; 
Customer Review and Rating Format  }\\
\cmidrule{2-3}
& Manufacturing Process Optimization     & \parbox[t]{\linewidth}{
Manufacturing Reports Format (MRF); 
Bill of Materials (BOM); 
Work Instruction Format (WIF); 
Standard Operating Procedure (SOP); 
Production Scheduling Format (PSF)   }       \\
\multirow{8}{*}{\shortstack[l]{Commerce\\
and \\
Manufact- \\ uring}}  & Inventory and Supply Chain Management  & \parbox[t]{\linewidth}{
Purchase Order (PO) Format; 
Inventory Report Format; 
Sales Forecast Format; 
Shipping Status Format; 
Return/Replacement Order Format }   \\
\cmidrule{2-3}
& Quality Control                        & \parbox[t]{\linewidth}{
Product Inspection Report; 
Quality Assurance (QA) Test Report; 
Defect Tracking Log; 
Product Compliance Certificate; 
Supplier Quality Report  } \\
\cmidrule{2-3}
& Smart Logistics and Route Optimization & \parbox[t]{\linewidth}{
Freight Bill Format; 
Inventory Update Format; 
Shipping Manifest Format; 
Route Optimization Report; 
Order Pick List } \\ 
\midrule
\multirow{13}{*}{\shortstack[l]{Customer \\ Relationship \\ Management \\ (CRM)}} & Customer Service   & \parbox[t]{\linewidth}{
Customer Email Response Format; 
Live Chat Transcript Format; 
Customer Feedback Form Response Format;
Ticketing System Response Format; 
Social Media Comment Response Format} \\
\cmidrule{2-3}
& Sales Forecasting                      & \parbox[t]{\linewidth}{
Sales Forecast Report; 
Key Performance Indicator (KPI) Report; 
Pipeline Report; 
Sales Targets Report; 
CRM Dashboard Summary }\\
\cmidrule{2-3}
& Recruitment Assistants                 &\parbox[t]{\linewidth}{
Resume Format; 
Job Description Format; 
Interview Schedule Format; 
Applicant Status Update Format; 
Candidate Comparative Analysis    }                                        \\
\cmidrule{2-3}
& Project Management                     & \parbox[t]{\linewidth}{
Gantt Chart Representation; 
Task Breakdown Structure (TBS); 
Project Status Reports; 
Meeting Minutes; 
Risk Assessment Reports       }                                         \\
\cmidrule{2-3}
 & Lead Scoring                           & \parbox[t]{\linewidth}{
 Lead Score Reports; 
 Sales Funnel Analysis Documents; 
 Lead Demographic Profiles; 
 Customer Interaction Logs; 
 Predictive Scoring Reports    }                                   \\ \midrule
\multirow{13}{*}{Marketing}                              & Consumer Behavior Analysis             & \parbox[t]{\linewidth}{
Consumer Behavior Report; 
Marketing Performance Dashboard; 
Advertising Copy Feedback; 
Social Media Sentiment Analysis; 
Competitor Analysis Summary    }                      \\
\cmidrule{2-3}
& Advertising Campaign Optimization      & \parbox[t]{\linewidth}{
Advertising Audience Profile Format; 
KPI Report Format;
A/B Test Result Format;
Competitive Analysis Format;
Campaign Budget Format    }  \\
\cmidrule{2-3}
& Content Curation and Creation          & \parbox[t]{\linewidth}{
Blog Post;
Social Media Post; 
Email Newsletters; 
Press Release; 
SEO Content     }  \\
\cmidrule{2-3}
 & Social Media Management                & \parbox[t]{\linewidth}{
 Social Media Report; 
 Content Calendar; 
 User Sentiment Analysis; 
 Hashtags Usage Report; 
 Social Media Customer Inquiries Response }   \\
 \cmidrule{2-3}
& Search Engine Optimization  & \parbox[t]{\linewidth}{
SERP (Search Engine Results Page) Report; 
SEO Keyword Analysis; 
On-Page SEO Audits; 
Backlink Profile Report; 
Competitor Analysis Report   }    \\ 
\midrule
\multirow{13}{*}{\shortstack[l]{Scientific \\
Research
and \\
Development}} & Mathematical Research                     & \parbox[t]{\linewidth}{
LaTeX;\\
MathML (Mathematical Markup Language);
SageMath Notebooks; 
Maple; 
MATLAB scripts    }\\
\cmidrule{2-3}
 & Physics   & \parbox[t]{\linewidth}{
 LaTeX (.tex); 
 MathML (.mathml); 
 BibTeX (.bib); 
 Research Paper Abstract Structured Text; 
 Physical Quantities and Units in UCUM (.ucum) }\\
 \cmidrule{2-3}
& Chemistry and Biological Sciences         & \parbox[t]{\linewidth}{
FASTA Format; 
PDB Format (Protein Data Bank);
GenBank Format; 
SMILES Format (Simplified Molecular Input Line Entry System); 
MOL and SDF Formats (MOLecular and Structure-Data File)}        \\
\cmidrule{2-3}
 & Environmental Sciences and Climate Change & \parbox[t]{\linewidth}{
 Research Paper (APA Format); 
 Scientific Report Format; 
 Environmental Impact Statement (EIS); 
 Peer Review Reports;
 Policy Briefs    }\\
 & Space Exploration                         & \parbox[t]{\linewidth}{
 NASA Planetary Data System (PDS) Format; 
 MISSION OPERATION REPORT (MOR) Format; 
 OBSErVation Time Series (OBSErVTS) Format; 
 Spacecraft Event Language (SEL) Format; 
 Telescope Observation Request (TOR) Format }              \\ 
 \midrule
\multirow{15}{*}{Education}                           & Adaptive Learning Platforms               & \parbox[t]{\linewidth}{
Personalized Learning Plan (PLP); 
Assessment Results Format (ARF); 
Interactive Course Content Format (ICCF);
Collaboration Log Format (CLF); 
Task Progress Report Format (TPRF)  }                                           \\
\cmidrule{2-3}
  & Intelligent Tutoring Systems              & \parbox[t]{\linewidth}{
  Lesson Summary Format; 
  Student Performance Report Format; 
  Quiz/Instruction Format; 
  Feedback/Correction Format; 
  Personalized Learning Path Recommendation Format  } \\
  \cmidrule{2-3}
   & Automated Grading Systems                 & \parbox[t]{\linewidth}{
   Rubric Score Format; 
   Student Report Format; 
   Class Rank Format; 
   Question Assessment Format; 
   Error Analysis Format  } \\
   \cmidrule{2-3}
  & Education Data Analysis                   & \parbox[t]{\linewidth}{
  Report in Academic Results Format (ARF);
  Student Behavior Analysis Format (SBAF);
  Educational Content Analysis Format (ECAF);
  Learning Style Analysis Format (LSAF);
  Teaching Performance Evaluation Format  }       \\
  \cmidrule{2-3}
 & Language Learning                         & \parbox[t]{\linewidth}{
 Learning Material Format; 
 Quizzes/Test Format; 
 Progress Report Format; 
 Language Translation Format; 
 Phonetic Script Format     }   \\ 
 \midrule
\multirow{12}{*}{Legal}                               & Contract Review and Analysis              & \parbox[t]{\linewidth}{
Legal Brief Format; 
Contract Abstract Format; 
Risk Assessment Format; 
Clause Breakdown Format; 
Legal Opinion Format }\\
 \cmidrule{2-3}
 & Legal Document Automation                 & \parbox[t]{\linewidth}{
 Legal XML (LegalXML); 
 Interactive Legal Applications Markup Language (iLAML)}                                                       \\ \cmidrule{2-3}
& Legal Research                            & \parbox[t]{\linewidth}{
Legal Brief; 
Case Citation; 
Contract Format; 
Statute    }  \\
\cmidrule{2-3}
 & Predictive Legal Analytics                & \parbox[t]{\linewidth}{
 Legal Reporting Document; 
 Case Brief Format; 
 Legal Opinion Letter Format; 
 Legal Case Study Format          }                                                                                        \\
 \cmidrule{2-3}
 & Intellectual Property (IP) Management     & \parbox[t]{\linewidth}{
 Patent Disclosure Forms; 
 Trademark Registration Documents;
 Copyright Registration Forms; 
 Intellectual Property Agreement Contracts; 
 Patent Litigation Documents       }   \\ 
 \midrule
\multirow{13}{*}{\shortstack[l]{Arts \\
and \\
Entertainment}}              & Music                                     & \parbox[t]{\linewidth}{
Lyrics Text Format; 
Chord Sheet Format; 
Tracklist Format; 
Metadata Format }   \\
\cmidrule{2-3}
 & Film Scriptwriting                        & \parbox[t]{\linewidth}{
 Screenplay Format; 
 Synopsis/Outline Format; 
 Treatment Format; 
 Beat Sheet Format; 
 Character Profile/Backstory Format}   \\
 \cmidrule{2-3}
 & Visual Art Creation                       & \parbox[t]{\linewidth}{
 SVG (Scalable Vector Graphics); 
 GLSL (OpenGL Shading Language);
 TikZ (code for creating vector graphics);
 POV-Ray Scene Description Language (SDL); 
 G-code (a language in which people instruct computerized machine tools) }\\
 \cmidrule{2-3}
 & Video Game Development                    & \parbox[t]{\linewidth}{
 Game Design Document (GDD); 
 Interactive Fiction Markup Language (IFML);
 Lua table for game configuration; 
 GLSL Shader Code; 
 Unreal Engine Blueprints Visual Scripting (Print String Node)    }                               \\
 \cmidrule{2-3}
& Sports Analytics and Performance          & \parbox[t]{\linewidth}{
Game Statistics Report; 
Training Performance Summary;
Player Ranking Report;
Injury Report; 
Match Prediction Report        }     \\ 
\end{longtable}
\twocolumn

\label{app:eval_temp}

\begin{figure*}[tp]
        \centering
        \begin{tabular}{|p{\linewidth}|}
        \hline
    \textbf{Evaluate Prompt Template}
    \\ \hline 
    <|im\_start|>system\\ You are a helpful assistant who evaluates the correctness and quality of models' outputs. \\
    <|im\_end|> \\
    <|im\_start|>user \\ I would like you to create a leaderboard that evaluates the correctness of the format of answers from various large language models. To accomplish this, you will need to analyze the text prompts given to the models and their corresponding answers. Specifically, please ensure that your evaluation outputs are properly formatted as a json string. I will provide both the prompts and the responses for this purpose. \\
    \\ 
    
    Here is the prompt: \\
    \{ \\
    \qquad "instruction": """\{instruction\}""", \\
    \}  \\ \\
    Here are the outputs of the models: \\
    {[}\\ 
        \qquad\{\\
            \qquad\qquad"model": "model",\\
            \qquad\qquad "answer": """\{output\}"""\\
        \qquad\},\\
    {]}\\ \\
    
    Please evaluate the formatting of the model's responses by checking if they comply with the format specifications stated in the prompt. Perform a thorough format check and provide a detailed explanation for why the format is correct or incorrect. Your feedback should include the name of the model, followed by the format correctness status represented as '1' for correct and '0' for incorrect. Present your reasoning as bullet points within a single string for each model assessed. In other words, you should produce the following output:\\
    ```json\\
    {[}\\
        \qquad\{\\
            \qquad\qquad"model": \textless{}model-name\textgreater{},\\
            \qquad\qquad"format\_correctness": \textless{}correctness\textgreater{},\\
        \qquad\qquad"reasons": \textless{}reasons-of-format-correctness\textgreater\\
    \qquad\}\\
    {]}'''\\ 
    \\
    Please note that your response should be a properly formatted JSON string and should not contain any additional content. We will load it directly as a JSON string in Python. \\
    <|im\_end|>\\
    \hline
        \end{tabular}
        \caption{Evaluate template prompt.}
        \label{fig:eval_temp}
\end{figure*}

\end{document}